\pdfoutput=1

\documentclass[11pt]{article}

\usepackage{acl}

\usepackage{times}
\usepackage{latexsym}
\usepackage{graphicx}
\usepackage{comment}
\usepackage{amsmath,amssymb}
\usepackage{bbding}
\usepackage{booktabs}
\usepackage{multirow}
\usepackage{colortbl}
\usepackage[whole]{bxcjkjatype} 
\usepackage{autobreak}
\usepackage{stfloats}

\newcolumntype{P}{>{\centering\arraybackslash}p{35.0pt}}

\usepackage[T1]{fontenc}
\usepackage[utf8]{inputenc}

\usepackage[utf8]{inputenc}

\usepackage{microtype}

\title{A Simple and Strong Baseline for End-to-End Neural RST-style\\ Discourse Parsing}
\newcommand{\mail}[2]{\href{mailto:#1}{\color{black}{#2}}}
\author{
    \textbf{Naoki Kobayashi}\textsuperscript{\rm 1},
    \textbf{Tsutomu Hirao}\textsuperscript{\rm 2},
    \textbf{Hidetaka Kamigaito}\textsuperscript{\rm 1},\\
    \textbf{Manabu Okumura}\textsuperscript{\rm 1}\textbf{,}
    \textbf{Masaaki Nagata}\textsuperscript{\rm 2}\\

    \textsuperscript{\rm 1}Institute of Innovative Research, Tokyo Institute of Technology,\\
    \textsuperscript{\rm 2}NTT Communication Science Laboratories, NTT Corporation\\
    
    \texttt{\{\mail{kobayasi@lr.pi.titech.ac.jp}{kobayasi@lr.},
              \mail{kamigaito@lr.pi.titech.ac.jp}{kamigaito@lr.},
              \mail{oku@pi.titech.ac.jp}{oku@}%
            \}pi.titech.ac.jp}\\
    \texttt{\{\mail{tsutomu.hirao.kp@hco.ntt.co.jp}{tsutomu.hirao.kp}, 
              \mail{masaaki.nagata.et@hco.ntt.co.jp}{masaaki.nagata.et}%
            \}@hco.ntt.co.jp}\\
}

\begin{document}
    \maketitle
    \begin{abstract}
To promote and further develop RST-style discourse parsing models, we need a strong baseline that can be regarded as a reference for reporting reliable experimental results. 
This paper explores a strong baseline by integrating existing simple parsing strategies, top-down and bottom-up, with various transformer-based pre-trained language models.
The experimental results obtained from two benchmark datasets
demonstrate that 
the parsing performance strongly relies on
the pre-trained language models rather than the parsing strategies.
In particular, the bottom-up parser achieves large performance gains compared to the current best parser when employing DeBERTa.
We further reveal that language models with a span-masking scheme especially boost the parsing performance through our analysis within intra- and multi-sentential parsing, and nuclearity prediction.
\end{abstract}

    \section{Introduction}

\begin{figure}[t]
    \centering
    \includegraphics[width=0.9\linewidth]{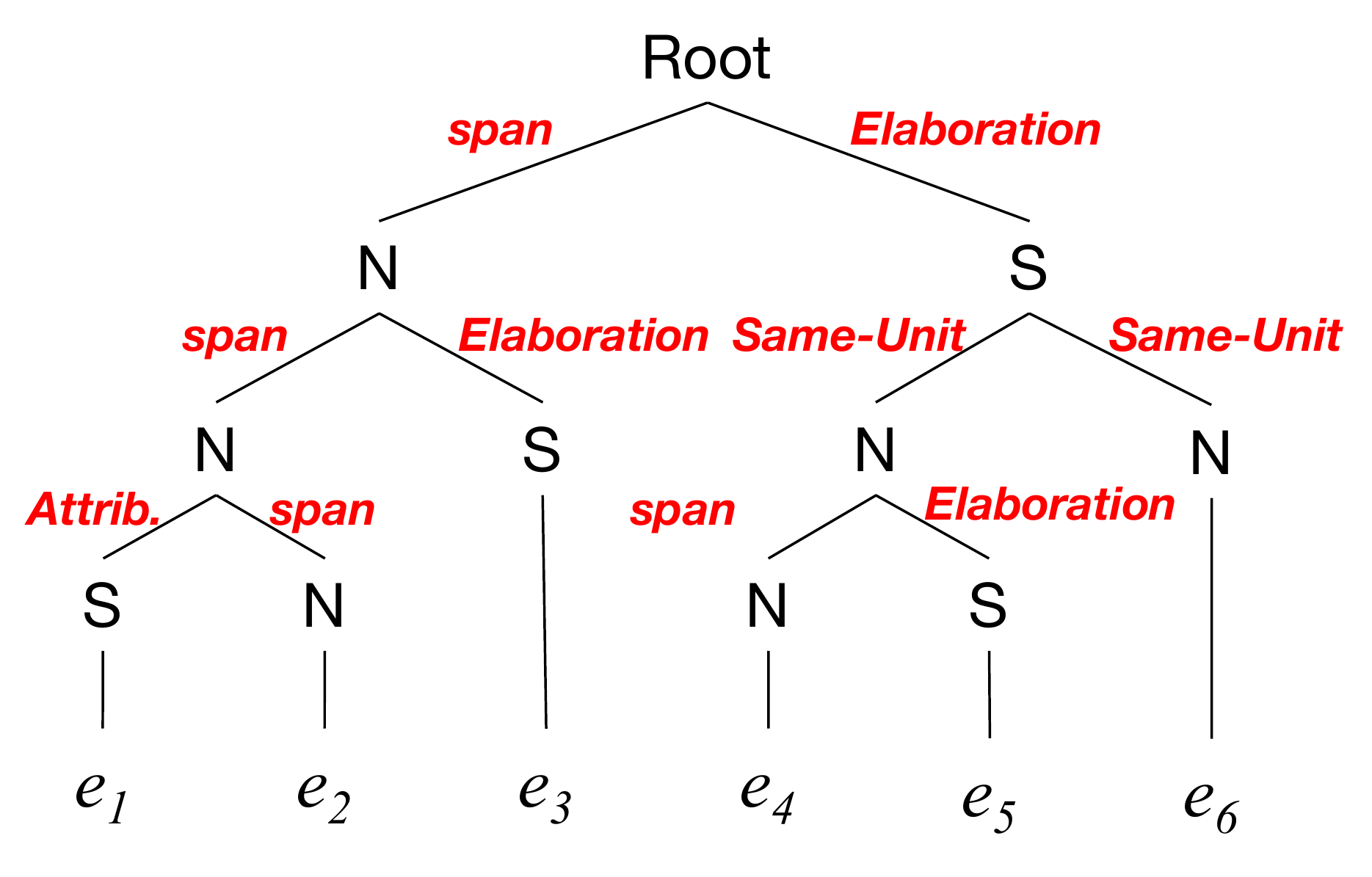}
    \caption{
        Example RST-style discourse tree, obtained from {\tt WSJ\_1100} in the RST discourse treebank \cite{carlson-etal-2002-rstdt}, consisting of six EDUs:
                $e_1$:[Westinghouse Electric Corp. said],
        $e_2$:[it will buy Shaw-Walker Co.],
        $e_3$:[Terms weren't disclosed.],
        $e_4$:[Shaw-Walker,],
        $e_5$:[based in Muskegon, Mich.,],
        $e_6$:[makes metal files and desks, and seating and office systems furniture.]. 
    }
    \label{fig:RSTtree}
\end{figure}

Rhetorical Structure Theory (RST)~\cite{mann:87:a} is one of the most influential theories for representing the discourse structure behind a document. According to the theory, a document is represented as a recursive constituent tree that indicates the relation between text spans consisting of a single elementary discourse unit (EDU) or contiguous EDUs. The label of a non-terminal node describes the nuclearity status, either nucleus or satellite, of the text span, and the edge indicates the rhetorical relation between the text spans
(Figure \ref{fig:RSTtree}). 

RST-style discourse parsing (hereafter RST parsing) is a fundamental task in NLP and plays an essential role in several downstream tasks, such as text summarization~\cite{liu-chen-2019-exploiting, xu-etal-2020-discourse}, question-answering~\cite{gao-etal-2020-discern}, and sentiment analysis~\cite{bhatia-etal-2015-better}. In most cases, the performance of an RST parsing method has been evaluated on the largest English treebank, the RST discourse treebank (RST-DT)~\cite{carlson-etal-2002-rstdt}, as the benchmark dataset. The evaluation measures used include the micro-averaged F1 score of unlabeled spans, that of nuclearity-labeled spans, that of rhetorical relation-labeled spans, and that of fully labeled spans, based on standard Parseval~\cite{morey-etal-2017-much}, when using gold EDU segmentation.

There are two major strategies for RST parsing: top-down and bottom-up. 
The former builds RST trees by splitting a larger text span consisting of EDUs into smaller ones recursively. The latter builds trees by merging two adjacent text spans.
Non-neural parsers with classical handcrafted features prefer the bottom-up strategy \cite{duverle-prendinger-2009-novel,feng-hirst-2012-text,wang-etal-2017-two}.
On the other hand, recent neural parsers prefer the top-down strategy \cite{kobayashi-etal-2020-topdown,koto-etal-2021-top,nguyen-etal-2021-rst,zhang-etal-2021-adversarial}, while a few parsers employ the bottom-up strategy \cite{guz-carenini-2020-coreference,guz-etal-2020-unleashing}.
With advances in neural network models, such neural parsers obtain significant gains over the non-neural parsers.

Several techniques have been proposed to boost RST parsing performance:
\citet{nguyen-etal-2021-rst} and \citet{shi2020endtoend} introduced beam search and  \citet{koto-etal-2021-top} exploited dynamic oracles in search to improve their parsing algorithms. 
\citet{kobayashi-etal-2020-topdown} used a model ensemble with multiple runs, 
\citet{zhang-etal-2021-adversarial} introduced adversarial training, and  \citet{kobayashi-etal-2021-improving} and \citet{guz-etal-2020-unleashing} exploited silver data to improve parameter optimization. 

Furthermore, pre-trained language models are playing an important role in improving the parsing performance.
\citet{shi2020endtoend}, \citet{nguyen-etal-2021-rst}, and \citet{zhang-etal-2021-adversarial} employed XLNet \cite{zhilin2019xlnet} to obtain better vector representations for arbitrary text spans consisting of EDU(s).
As a result, the current best top-down parser, \citet{zhang-etal-2021-adversarial} with XLNet, achieved the fully labeled span F1 score of 53.8 with standard Parseval.
The method has a gain of 4-5 points compared to the best non-neural parser \cite{wang-etal-2017-two}. However, it is still unclear what contributed the most to the improvement among the models' various factors such as parsing strategies, pre-trained language models, and a model ensemble.
Therefore, we need a simple but strong baseline for different parsing strategies along with a pre-trained language model to clarify the effects of the parsing strategies and pre-trained models.
The baseline will contribute to building more reliable experiments for revealing the effectiveness of newly proposed methods.\footnote{
Building strong baselines can be recently regarded as an essential issue for various NLP tasks \cite{liang-etal-2019-strong,suzuki-etal-2018-empirical,denkowski-neubig-2017-stronger}.
}

This paper aims to build strong baselines for RST parsing, based on two simple open-source top-down \cite{kobayashi-etal-2020-topdown} and bottom-up \cite{guz-carenini-2020-coreference} parsers, employing transformer-based pre-trained language models, without incorporating any of our own mechanisms.\footnote{
Note that we introduced some minor modifications, as we mention in Section~\ref{sec:parsing}.} 
The experimental results on RST-DT \cite{carlson-etal-2002-rstdt} and Instructional Discourse Treebank (Instr-DT) \cite{subba-di-eugenio-2009-effective} with various pre-trained language models
demonstrated that the parsing performance strongly relies on
the performance of the pre-trained language models rather than the parsing strategies.
While the current trend is a top-down parser, a bottom-up parser with DeBERTa \cite{he2021deberta}, one of the current state-of-the-art pre-trained language models, achieved the best score, which is higher than those of the current state-of-the-art parsers.
Further, our analysis based on intra- and multi-sentential parsing, and nuclearity prediction
revealed that pre-trained language models with a {\it span-masking} scheme improve parsing performance more than those with a {\it token-masking} scheme.
We will release our code at \url{https://github.com/nttcslab-nlp/RSTParser_EMNLP22}.

\renewcommand{\arraystretch}{0.85}
\begin{table*}[t]
\small
    \centering
    \begin{tabular}{lllcccc}
    \toprule
    \textbf{Model} & \textbf{Strategy} & \textbf{Language Model} &\textbf{Span} & \textbf{Nuc.} & \textbf{Rel.} & \textbf{Full} \\
    \midrule
Wang et al. \shortcite{wang-etal-2017-two} & Bottom-up & $-$ & 72.0 & 60.5 & 50.4 & 48.2\\
    \cmidrule{1-7}
    Guz and Carenini \shortcite{guz-carenini-2020-coreference} & Bottom-up  &SpanBERT & 75.8 & 64.6 & 53.7 & 51.4\\
    Guz et al. \shortcite{guz-etal-2020-unleashing} & Bottom-up & RoBERTa & 72.9 & 61.9 & $-$ & $-$\\
    Koto et al. \shortcite{koto-etal-2021-top} & Top-down & GloVe & 73.1 & 62.3 & 51.5 & 50.3\\
     Kobayashi et al. \shortcite{kobayashi-etal-2021-improving} & Top-down & Glove+ELMo & 74.1 & 64.7 & 54.1 & 52.7\\
    Nguyen et al. \shortcite{nguyen-etal-2021-rst}&Top-down & XLNet & 74.3 & 64.3 & 51.6 & 50.2\\
    Zhang et al. \shortcite{zhang-etal-2021-adversarial} & Top-down & XLNet & 76.3 & 65.5 & 55.6 & 53.8\\
    \bottomrule
    \end{tabular}
    \caption[Caption for LOF]{
    Comparison between the best non-neural parser and recent end-to-end neural parsers.
    }
    \label{tab:compare_sota_models}
\end{table*}

\section{Related Work}
Early studies on RST parsing were based on non-neural supervised learning methods with handcrafted features.
As parsing strategies,  bottom-up greedy algorithms \cite{duverle-prendinger-2009-novel,feng-hirst-2012-text},
 shift-reduce \cite{wang-etal-2017-two}, CRFs \cite{feng-hirst-2014-linear}, and CKY-like parsing algorithms \cite{joty-etal-2013-combining, joty-etal-2015-codra} have been employed.
In particular, Wang et al.'s shift-reduce parser \cite{wang-etal-2017-two}, based on SVMs, achieved 
the best results among the non-neural statistical models on the RST-DT. The method first builds nuclearity-labeled RST trees and then assigns relation labels between two adjacent spans consisting of a single or multiple EDUs.

Inspired by the success of neural networks in many NLP tasks, several early neural network-based models have been proposed for RST parsing \cite{ji-eisenstein-2014-representation, li-etal-2014-recursive, li-etal-2016-discourse}. 
However, as reported by \citet{morey-etal-2017-much}, while some neural approaches outperformed classical approaches, it was not by a large margin.

Recent end-to-end neural RST parsing models with sophisticated language models, such as GloVe and ELMo, achieved better performance.
They used vector representations of text spans based on the LSTMs whose inputs are 
word embeddings from the language models.
\citet{yu-etal-2018-transition} proposed a bottom-up parser, based on the shift-reduce algorithm, 
that leverages the information from their neural dependency parsing model within a sentence for RST parsing.
The method outperformed traditional non-neural methods and obtained a remarkable 
relation-labeled span F1 score of 49.2.
As another approach,
a top-down neural parser based on a sequence-to-sequence (seq2seq) framework was proposed~\cite{lin-etal-2019-unified} for use only at the sentence level.
The method parses a tree in a depth-first manner with a pointer-generator network.
\citet{zhang-etal-2020-top} extended the method and applied it to document-level RST parsing.
\citet{kobayashi-etal-2020-topdown} proposed another top-down RST parsing method, based on a minimal span-based approach, that splits a span into smaller ones recursively and exploits multiple granularity levels in a document.
Then, they demonstrated the impact of the model ensemble.
\citet{koto-etal-2021-top} extended Kobayashi et al.'s parser \shortcite{kobayashi-etal-2020-topdown} by introducing dynamic oracles as well as a new penalty for segmentation loss, which is based on the current tree depth and the number of EDUs in the input.
The latter two methods also outperformed traditional non-neural methods.

More recently, neural RST parsing models with transformer-based pre-trained language models,
such as SpanBERT and XLNet, have been proposed.
\citet{guz-carenini-2020-coreference} extended Wang et al.'s bottom-up parser \shortcite{wang-etal-2017-two} 
by replacing SVMs with a neural classifier and employing SpanBERT to obtain representations for text spans.
The performance was greatly improved: They achieved nuclearity-labeled and relation-labeled span F1 scores of 64.6 and 51.4, respectively.
\citet{shi2020endtoend} extended Lin et al.'s top-down model \shortcite{lin-etal-2019-unified} 
by introducing layer-wise beam search and XLNet.
\citet{nguyen-etal-2021-rst} also introduced beam search in their seq2seq-based top-down model and reported
that XLNet greatly contributed to improving performance.
\citet{zhang-etal-2021-adversarial} improved a seq2seq-based top-down model by exploiting adversarial training.
They also reported that performance was boosted by XLNet, and the following current best scores were obtained: 76.3, 65.5, 55.6, and 53.8 
for unlabeled, nuclearity-, relation-, and fully labeled span F1 
scores, respectively.

As another approach, 
\citet{guz-etal-2020-unleashing} and \citet{kobayashi-etal-2021-improving} 
proposed a pre-training and fine-tuning framework for RST parsing. They obtained silver data from
automatically parsed large-scale data and used them to pre-train their models. Then, they fine-tuned the models with gold data.

Table~\ref{tab:compare_sota_models} summarizes the previous best non-neural parser and recent end-to-end neural RST parsers with performance that can be considered state-of-the-art.
We can see that the RST parsing models with the transformer-based language models 
outperformed the other models regardless of the parsing strategy. 
The performance improvements are remarkable, especially for the relation-labeled and fully labeled span F1 scores.

    \section{End-to-end Neural RST Parsing \label{sec:parsing}}
\begin{figure}[t]
    \centering
    \includegraphics[width=0.75\linewidth]{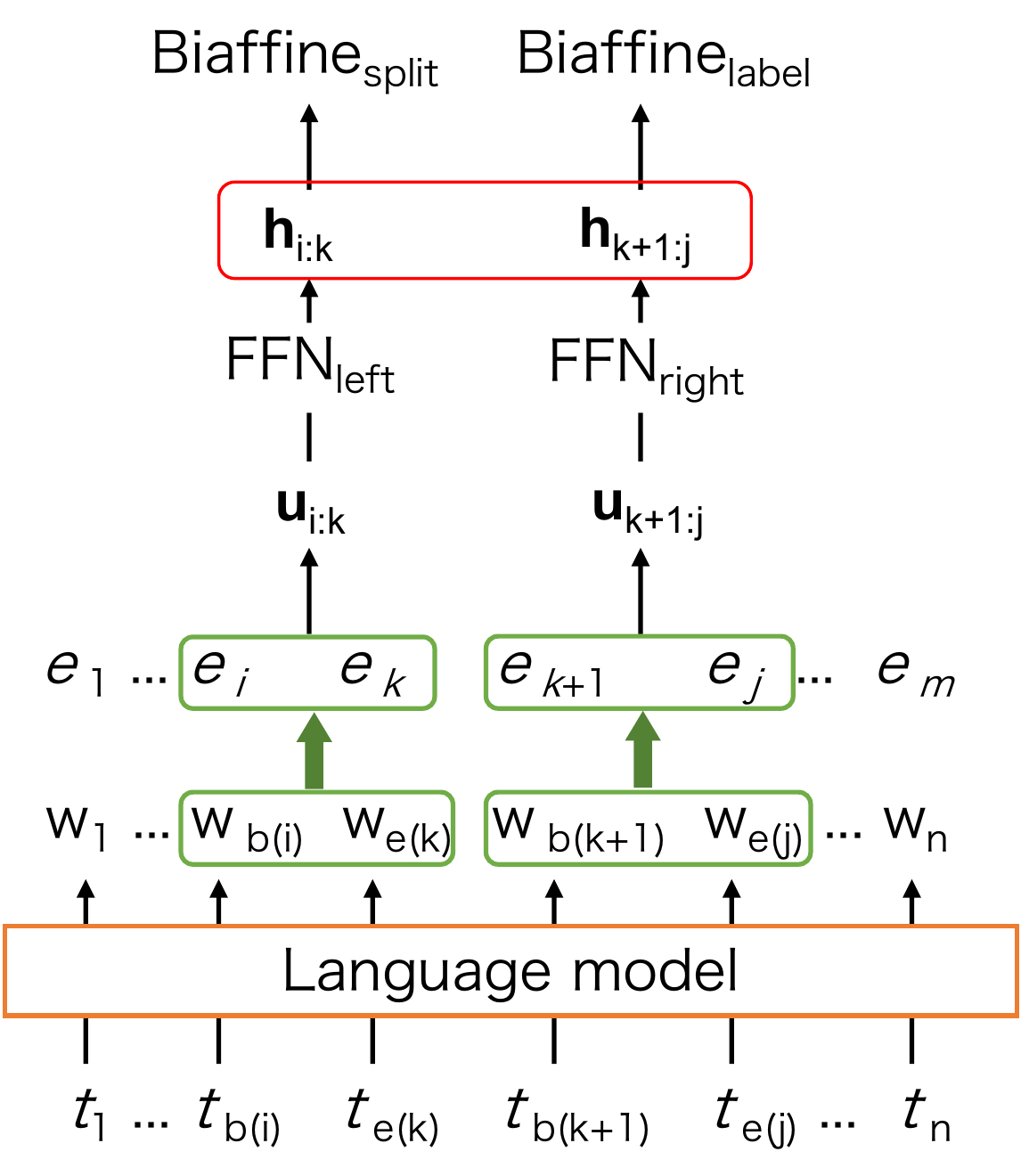}
    \caption{Top-down parsing model.}
    \label{fig:topdown}
\end{figure}

\begin{figure}[tb]
    \begin{center}
        \includegraphics[clip,height=0.8\linewidth,width=0.95\linewidth]{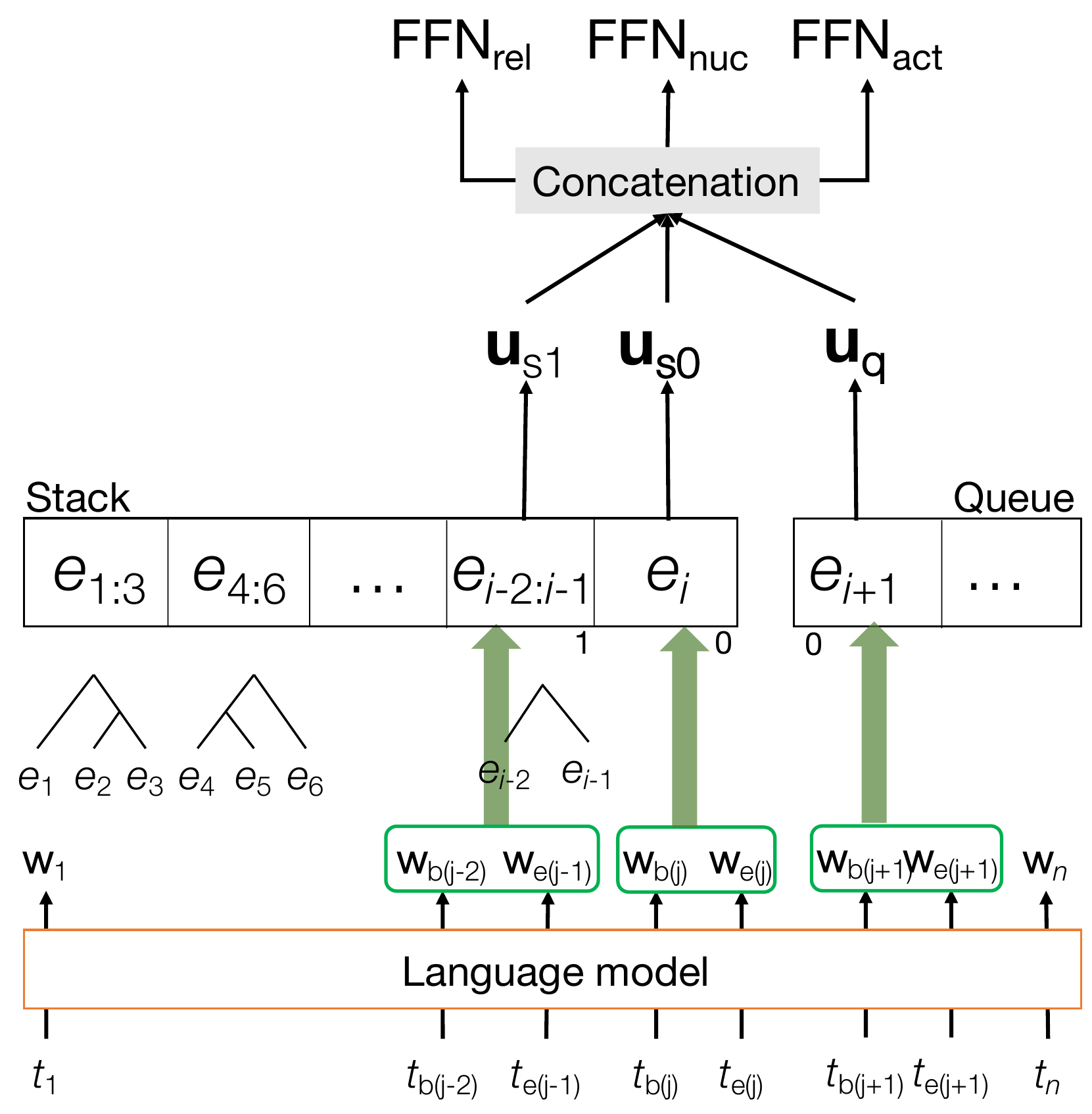}
        \caption{Bottom-up parsing model.}
        \label{fig:bottomup}
    \end{center}
\end{figure}

We employed the span-based  parser~\cite{kobayashi-etal-2020-topdown,koto-etal-2021-top} for the top-down parsing strategy and the shift-reduce transition parser \cite{guz-carenini-2020-coreference}, an end-to-end variant of Wang et al.'s parser  \shortcite{wang-etal-2017-two}, for the bottom-up parsing strategy.
These parsers were chosen here for their simple architecture and their open code.
Overviews of the parsers are shown in Figs.~\ref{fig:topdown} and \ref{fig:bottomup}.
Both parsers basically consist of simple Feed-Forward Networks (FFNs) and BERT-based embeddings.
In this study, we used a two-layer perceptron with the GELU activation function and a dropout layer.

\subsection{Vector Representations for Text Spans}
Before describing the parsing models, we explain how to obtain a vector
representation for an arbitrary text span by using BERT-based language models. 
Our procedure for obtaining the vector representation is a simplified variant of Guz and Carenini's method~\shortcite{guz-carenini-2020-coreference}.

First, we transform a document into a sequence of subwords, $\{t_1,t_2,\ldots,t_n\}$. 
Then, we obtain the vector representation for each subword in the sequence $\{\mathbf{w}_1,\mathbf{w}_2,
\ldots,\mathbf{w}_n\}$ using a language model.
Following~\citet{guz-carenini-2020-coreference}, the vector representation of a text span $\mathbf{u}_{i:j}$, consisting of the $i$-th EDU to the $j$-th EDU, is obtained by averaging the vector of both edge subwords, $\mathbf{u}_{i:j}=(\mathbf{w}_{\text{b}(i)}+\mathbf{w}_{\text{e}(j)})/2$, where b($i$) returns the index of the leftmost subword in the $i$-th EDU and 
e($j$) returns that of the rightmost subword in the $j$-th EDU.
A document longer than the maximum allowed length of BERT (512 subwords) is embedded with sliding windows with 30-subword over-wrapping. 

\subsection{Top-down Parsing}

Since the minimal span-based parser does not require any additional module like a decoder, as in the pointer-network-based top-down parsers, 
it is suitable for a comparison with a bottom-up parser of the shift-reduce algorithm, which consists of three simple FFNs, as we describe in Section~\ref{seq:BU_parsing}.
This top-down parser splits each span into smaller ones recursively until the span becomes a single EDU. 
We modified 
the code\footnote{
\url{https://github.com/nttcslab-nlp/Top-Down-RST-Parser}
} to utilize transformer-based embeddings 
and simplified it by excluding organizational features that represent sentence and paragraph boundary information.
By following \citet{koto-etal-2021-top}, we introduced a biaffine layer for span splitting and a loss penalty.

For each position $k$ in a span consisting of the $i$-th EDU to the $j$-th EDU, a scoring function, $s_{split}(i, j, k)$, is defined as follows:
\begin{align}
    \begin{autobreak}
    s_{\text{split}}(i, j, k) = 
        \mathbf{h}_{i:k}\mathbf{W} \mathbf{h}_{k+1:j}
        + \mathbf{v}_{\rm left} \mathbf{h}_{i:k}
        + \mathbf{v}_{\rm right} \mathbf{h}_{k+1:j}, 
    \end{autobreak}
     \label{split_score}
\end{align}
where $\mathbf{W}$ is a weight matrix and $\mathbf{v}_{\rm left}$, $\mathbf{v}_{\rm right}$ are weight vectors corresponding to the left and right spans, respectively.
Here, $\mathbf{h}_{i:k}$ and  $\mathbf{h}_{k+1:j}$
are defined as follows:
\begin{align}
    \mathbf{h}_{i:k} &= \textrm{FFN}_{\text{left}}(\mathbf{u}_{i:k}), \\
    \mathbf{h}_{k{+}1:j} &= \textrm{FFN}_{\text{right}}(\mathbf{u}_{k{+}1:j}), 
    \label{eq:td_split}
\end{align}
Then, the span is split at position $k$ that maximizes Eq.~(\ref{split_score}):
\begin{equation}
    \hat{k}=\mathop{\rm argmax}_{i \le k < j} s_{\text{split}}(i,j,k).
    \label{eq:td_estimate_split}
\end{equation}
When splitting a span at position $k$, the score of the nuclearity status and relation labels for the two spans is defined as follows:
\begin{align}
    \begin{autobreak}
    s_{\text{label}}(i, j, \hat{k},\ell) = 
        \mathbf{h}_{i:\hat{k}}\mathbf{W}^{\ell} \mathbf{h}_{\hat{k}+1:j}
        {+} \mathbf{v}_{\rm left}^{\ell} \mathbf{h}_{i:\hat{k}}
        {+} \mathbf{v}_{\rm right}^{\ell} \mathbf{h}_{\hat{k}+1:j},  
    \end{autobreak}
     \label{label_score}
\end{align}
where $\mathbf{W}^{\ell}$ is a weight matrix for a specific label $\ell$, and 
 $\mathbf{v}_{\rm left}^{\ell}$ and $\mathbf{v}_{\rm right}^{\ell}$
 are weight vectors corresponding to the left and right spans for the label $\ell$, respectively.
While the correct split position in the training data is used for $\hat{k}$ in training time, the position predicted with Eq.~(\ref{eq:td_estimate_split}) is used in testing time.

Then, the label that maximizes Eq.~(\ref{label_score}) is assigned to the spans:
\begin{equation}
    \hat{\ell} = \mathop{\mathrm{argmax}}_{\ell \in {\cal L}} s_{\rm label}(i,j,\hat{k},\ell), \label{eq:td_estimate_label}
\end{equation}
where ${\cal L}$ denotes a set of valid nuclearity status combinations, \{N-S,S-N,N-N\}, for predicting the nuclearity 
and a set of relation labels, \{Elaboration, Condition,$\ldots$\}, 
for predicting the relation.
Note that the weight parameters $\mathbf{W}^{\ell}$ and $\text{FFNs}$ for the nuclearity and relation labeling are learned separately.\footnote{\citet{koto-etal-2021-top} reported that jointly predicting nuclearity and relation labels would improve performance. However, this might be due to the high-frequency label bias, and the total performance degraded in macro averaging. Thus, we predict them independently.}

\subsection{Bottom-up Parsing}\label{seq:BU_parsing}

Formally, in shift-reduce parsing, a parsing state is denoted as a tuple $(S, Q)$, where $S$ is a stack
 and $Q$ is a queue that contains incoming EDUs.
Each element in $S$ can be a completed constituent or a terminal.
At each step, the parser chooses one of the following actions with a neural classifier and updates the state:
\begin{itemize}
    \item \textsc{Shift}: pop the first EDU off the queue and push it onto the stack.
    \item \textsc{Reduce}: pop two elements from the stack and push a new constituent that has the popped subtrees as its children onto the stack as a single composite item.
\end{itemize}
In the \textsc{Reduce} action, nuclearity status and relation labels are predicted by different neural classifiers. 
That is, RST trees are built in three stages: First, unlabeled trees are built and then nuclearity status and relation labels are assigned independently. 
Note that previous shift-reduce parsers were based on the two-stage approach, which means they first build nuclearity-labeled RST trees and then assign relation labels to the trees.
To fairly compare top-down and bottom-up approaches, we employed the three-stage approach both in top-down and bottom-up parsing.
Our experiments demonstrated that there is no significant difference between the performances of the 
two-stage and three-stage approaches.

Our bottom-up parser has three classifiers, $\textrm{FFN}_{\rm act}$, $\textrm{FFN}_{\rm nuc}$, and $\textrm{FFN}_{\rm rel}$, for predicting action, nuclearity, and rhetorical relations, respectively. The difference among them is only the output dimension related to the number of classes; specifically, the output dimension of the action classifier is two (shift or reduce), that of the nuclearity classifier is three (N-S, N-N, S-N), and that of the relation classifier is the number of the rhetorical relations utilized in the dataset.
\begin{equation}
    s_{\rm *} {=} \textrm{FFN}_{\rm *}(
        \textrm{Concat}(\mathbf{u}_{{\rm s}_0}, \mathbf{u}_{{\rm s}_1}, \mathbf{u}_{{\rm q}_0})),
\end{equation}
where function ``Concat'' concatenates the vectors received as the arguments.
$\mathbf{u}_{{\rm s}_0}$ is the vector representation of a text span stored in
the first position of the stack $S$,
$\mathbf{u}_{{\rm s}_1}$ is that in the second position of $S$,
and $\mathbf{u}_{{\rm q}_0}$ is that in the first position of the queue $Q$.
Weights for each FFN and the language model used for span embeddings are trained by optimizing the cross-entropy loss of $s_{\rm act}$, $s_{\rm nuc}$, and $s_{\rm rel}$.
Note that we do not utilize organizational features as in top-down parsing.
    \section{Pre-trained Language Models}
\renewcommand{\arraystretch}{0.8}
\begin{table}[t]
\small
    \centering
    \begin{tabular}{lrl}
        \toprule
        LM & Data Size & Objective \\
        \midrule
        BERT            & 16GB          & MLM+NSP\\  
        RoBERTa         & 161GB         & MLM\\  
        XLNet           & 161GB         & PLM\\  
        SpanBERT        & 16GB          & MLM\\  
        DeBERTa         & 85GB          & MLM\\  
        \bottomrule
    \end{tabular}
    \caption{Data size and task for pre-training. The data size is from \cite{he2021deberta}.
    }
    \label{tab:data_size_of_LM}
\end{table}

Since most of the transformer-based pre-trained language models originated in BERT, 
we employed BERT and four of its variants as language models to obtain vector representations
for text spans.
Table~\ref{tab:data_size_of_LM} shows the size of the dataset and the tasks for their pre-training.

\noindent {\bf BERT:} is trained with two tasks: (1) a masked language model (MLM); 15\% of the tokens in the training data are randomly masked, and then the model is trained to predict the
masked tokens, and (2) a next sentence prediction (NSP) task; 
the model is trained to correctly predict the following sentence for a given sentence.
BERT is trained on Book Corpus and English Wikipedia.

\noindent {\bf RoBERTa:} 
is trained with longer and larger batches over more data and longer sequences. 
It further removes NSP and 
dynamically changes the masking pattern applied to the training data. 
In addition to the dataset used for training BERT,
the training dataset here includes CC-News, OpenWebText, and Stories as well.

\noindent {\bf XLNet:} is a generalized autoregressive 
pre-trained language model, known as a permuted language model (PLM).
It is trained by maximizing the expected likelihood over all permutations of the factorization order of the input text to approximately consider bidirectional context.
In addition to the dataset used for training BERT,
the training dataset includes Giga5, ClueWeb, and CC as well.

\noindent {\bf SpanBERT:} is trained with
(1) a masked language model with random spans (contiguous tokens)
and (2) span boundary token prediction in the masked span.
Unlike the original BERT, SpanBERT does not include the NSP task.
The dataset used for training is the same as that for BERT.

\noindent {\bf DeBERTa:} is a modified variant of RoBERTa. It uses disentangled attention and an enhanced mask decoder.
During training, it masks spans randomly as for SpanBERT.
While the dataset used for training is a subset of that used for RoBERTa, 
it performs consistently better on various NLP tasks.

    \section{Experimental Settings}
\subsection{Datasets}
We used the RST-DT and Instr-DT to evaluate the performance of the parsers.
RST-DT contains 385 documents selected from the Wall Street Journal.
It is officially divided into 347 documents as the training dataset and 38 documents as the test dataset. 
The number of rhetorical relation labels utilized in the dataset is 18.
Since there is no official development dataset, we used 40 documents of the training dataset as the development dataset by following a previous study \cite{heilman2015fast}.
Instr-DT contains 176 documents of the home-repair instruction domain.
The number of rhetorical relation labels in the dataset is 39.
Since there are no official development and test datasets, we used 126,
25, and 25 documents for training, development, and test datasets, respectively.
We used gold EDU segmentation for both datasets by following conventional studies.

\subsection{Evaluation Metrics \label{seq:metrics}}
As in previous studies, 
we transformed RST-trees into right-heavy binary trees \cite{sagae-lavie-2005-classifier}
and evaluated the system results with micro-averaged F$_1$ scores of Span, Nuclearity, Relation, and Full, based on  Standard-Parseval \cite{morey-etal-2017-much}.
Span, Nuclearity, Relation, and Full were used to evaluate unlabeled, nuclearity-, relation-, and fully labeled tree structures, respectively.

Since neural models heavily rely on their given initialization, we report average scores and standard deviations of three runs with different seeds. 
\renewcommand{\arraystretch}{0.85}
\begin{table*}[t]
    \centering
    \small
    \begin{tabular}{llPPPPPPPP}
    \toprule
    & & \multicolumn{4}{c}{\textbf{RST-DT}} & \multicolumn{4}{c}{\textbf{Instr-DT}}\\       
    \cmidrule(lr){3-6}\cmidrule(lr){7-10}
    & & \textbf{Span} & \textbf{Nuc.} & \textbf{Rel.} & \textbf{Full} & \textbf{Span} & \textbf{Nuc.} & \textbf{Rel.} & \textbf{Full} \\
    \midrule
    \multicolumn{2}{l}{Zhang et al. } & 
                    76.3 & 65.5 & 55.6 & 53.8 &
                    $-$ & $-$ & $-$ & $-$\\
    \midrule 
    \multirow{5}{*}{\rotatebox[origin=c]{90}{Top-down}}
        & BERT      & 
                    69.8~(0.64)	& 59.1~(0.48)	& 48.3~(0.43)	& 46.6~(0.52) &
                    65.3~(0.81)	& 44.6~(1.32)	& 37.6~(0.38)	& 30.9~(0.33)\\
        & RoBERTa   &
                    77.3~(0.13)$^\clubsuit$	& 66.6~(0.18)$^\clubsuit$	& 55.8~(0.34)$^\clubsuit$	& 53.8~(0.37)$^\diamondsuit$ &
                    75.7~(0.45)$^\clubsuit$	& 56.1~(0.55)$^\clubsuit$	& 48.7~(0.63)$^\diamondsuit$	& 41.5~(0.47)$^\diamondsuit$\\
        & XLNet     & 
                    77.8~(0.16)$^\diamondsuit$	& 67.4~(0.05)$^\diamondsuit$	& \textbf{57.0~(0.38)}$^\diamondsuit$	& \textbf{54.8~(0.41)}$^\diamondsuit$ &
                    74.3~(0.13)$^\clubsuit$	& 55.2~(0.40)$^\clubsuit$	& 47.0~(0.57)$^\diamondsuit$	& 40.2~(0.51)$^\diamondsuit$ \\
        & SpanBERT  & 
                    76.5~(0.59)$^\clubsuit$	& 65.4~(0.37)$^\clubsuit$	& 54.5~(1.01)$^\clubsuit$	& 52.2~(0.84)$^\clubsuit$ &
                    73.7~(0.92)$^\clubsuit$	& 54.5~(0.88)$^\clubsuit$	& 42.7~(0.19)$^\clubsuit$	& 36.7~(0.22)$^\clubsuit$\\
        & DeBERTa   & 
                    \textbf{78.5~(0.14)}$^\heartsuit$	& \textbf{67.9~(0.23)}$^\diamondsuit$	& 56.6~(0.11)$^\diamondsuit$	& 54.4~(0.23)$^\diamondsuit$ &
                    \textbf{77.3~(0.66)}$^\spadesuit$	& \textbf{57.9~(0.47)}$^\clubsuit$	& \textbf{50.0~(0.38)}$^\spadesuit$	& \textbf{43.4~(0.26)}$^\diamondsuit$ \\
    \midrule
    \multirow{5}{*}{\rotatebox[origin=c]{90}{Bottom-up}} 
        & BERT      & 
                    68.3~(0.60)	& 57.8~(0.69)	& 47.8~(1.04)	& 46.0~(0.82) &
                    66.6~(1.04)	& 46.3~(1.63)	& 39.5~(1.59)	& 32.9~(1.20)\\
        & RoBERTa   & 
                    76.1~(0.33)$^\clubsuit$	& 66.5~(0.23)$^\clubsuit$	& 55.4~(0.38)$^\clubsuit$	& 53.7~(0.15)$^\clubsuit$ &
                    73.2~(0.85)$^\clubsuit$	& 55.5~(0.25)$^\clubsuit$	& 47.9~(0.70)$^\clubsuit$	& 41.4~(0.38)$^\clubsuit$\\
        & XLNet     & 
                    76.1~(0.67)$^\clubsuit$	& 65.9~(0.19)$^\clubsuit$	& 56.3~(0.55)$^\diamondsuit$	& 54.2~(0.48)$^\diamondsuit$ &
                    73.6~(1.31)$^\clubsuit$	& 56.4~(1.59)$^\clubsuit$	& 47.4~(0.47)$^\clubsuit$	& 40.7~(1.24)$^\clubsuit$\\
        & SpanBERT  & 
                    76.1~(0.37)$^\clubsuit$	& 65.3~(0.54)$^\clubsuit$	& 54.9~(0.13)$^\clubsuit$	& 52.7~(0.32)$^\clubsuit$ &
                    72.9~(1.26)$^\clubsuit$	& 53.8~(1.97)$^\clubsuit$	& 46.0~(1.28)$^\clubsuit$	& 40.5~(1.32)$^\clubsuit$\\
        & DeBERTa   & 
                    \textbf{77.8~(0.31)}$^\bigstar$	& \textbf{68.0~(0.48)}$^\bigstar$	& \textbf{57.3~(0.21)}$^\bigstar$	& \textbf{55.4~(0.36)}$^\bigstar$ &
                    \textbf{77.8~(0.59)}$^\bigstar$	& \textbf{60.0~(1.26)}$^\bigstar$	& \textbf{51.4~(1.40)}$^\bigstar$	& \textbf{44.4~(1.20)}$^\bigstar$\\
        \bottomrule
    \end{tabular}
    \caption{Results with various language models (\textbf{Standard-Parseval}). Standard deviations for three runs are shown in parentheses.
    The best score 
    is indicated in \textbf{bold}.
    $\bigstar$ indicates significantly better than any model except DeBERTa.  
    $\spadesuit$ indicates significantly better than BERT, XLNet, and SpanBERT. $\heartsuit$ indicates
    significantly better than BERT, RoBERTa, and SpanBERT. $\diamondsuit$ indicates significantly better than
    BERT and SpanBERT. $\clubsuit$ indicates significantly better than BERT.
    }
    \label{tab:result_base:horizontal_concat}
\end{table*}

\subsection{Training Configurations}
We implemented all models based on 
PyTorch~\cite{NEURIPS2019_9015} with 
PyTorch~Lightning~\cite{falcon2019pytorch} 
and used language models from Transformers~\cite{wolf-etal-2020-transformers}.
We used \texttt{base} models, such as \texttt{BERT-base-cased}, \texttt{RoBERTa-base}, for all the experiments.
The dimension of hidden layers in FFNs was set to 512, and the dropout rate was set to 0.2.
By following \citet{guz-carenini-2020-coreference}, we employed
span-based batch rather than document-based batch.
The mini-batch size is 5 spans/action.
We optimized all models with the AdamW~\cite{loshchilov2019decoupled} optimizer.
We used a learning rate of 1e-5 for language models and 1e-5/2e-4 for other parameters\footnote{The learning rate was determined by using the development dataset.} such as FFN and biaffine layers.
We scheduled the learning rate by linear warm-up, which increases the learning rate linearly during the first epoch and then decreases it linearly to 0 until the final epoch.
We trained the model up to 20 epochs and applied early stopping with a patience of 5 by monitoring the fully labeled span F1 score 
on the development dataset.
Details of other hyperparameters are in Appendix \ref{app:hyperparames}.

    \section{Results}
\renewcommand{\arraystretch}{0.85}
\begin{table*}[t]
    \centering
    \small
    \begin{tabular}{llPPPPPPPP}
    \toprule
    & & \multicolumn{4}{c}{\textbf{RST-DT}} & \multicolumn{4}{c}{\textbf{Instr-DT}}\\       
    \cmidrule(lr){3-6}\cmidrule(lr){7-10}
    & & \textbf{Span} & \textbf{Nuc.} & \textbf{Rel.} & \textbf{Full} & \textbf{Span} & \textbf{Nuc.} & \textbf{Rel.} & \textbf{Full} \\
    \midrule 
    \multirow{5}{*}{\rotatebox[origin=c]{90}{Top-down}}
        & BERT      & 
                    92.6~(0.53)	& 85.7~(0.41)	& 75.4~(0.45)	& 74.7~(0.54) &
                    82.7~(0.91)	& 69.1~(0.42)	& 52.7~(1.40)	& 47.4~(1.09)\\
        & RoBERTa   &
                    94.1~(0.46)	& 88.4~(0.46)	& 79.6~(0.17)	& 78.7~(0.11)  &                  
                    85.9~(0.82)	& 72.3~(1.14)	& 56.9~(1.00)	& 52.3~(1.08)\\                    
        & XLNet     & 
                    \textbf{94.8~(0.39)}	& \textbf{89.5~(0.39)}	& \textbf{80.5~(0.59)}	& \textbf{79.5~(0.53)}  &                  
                    85.8~(0.23)	& 73.2~(0.90)	& \textbf{58.4~(1.06)}	& \textbf{54.2~(1.43)}\\
        & SpanBERT  & 
                    94.1~(0.15)	& 88.8~(0.19)	& 79.4~(0.49)	& 78.5~(0.39) &
                    85.5~(0.68)	& 71.9~(0.69)	& 52.4~(1.33)	& 48.5~(1.60)\\
        & DeBERTa   & 
                    94.2~(0.33)	& 89.0~(0.16)	& 80.1~(0.43)	& 79.1~(0.32) &       
                    \textbf{86.9~(0.34)}	& \textbf{73.6~(0.41)}	& 58.2~(0.23)	& 53.8~(0.38)\\
    \midrule
    \multirow{5}{*}{\rotatebox[origin=c]{90}{Bottom-up}} 
        & BERT      & 
                    91.9~(0.34)	& 84.4~(0.31)	& 74.4~(0.37)	& 73.8~(0.30) &
                    84.5~(0.42)	& 69.4~(1.17)	& 53.7~(1.04)	& 48.8~(1.42)\\
        & RoBERTa   & 
                    94.4~(0.12)	& 89.0~(0.34)	& 80.4~(0.47)	& 79.7~(0.51) &
                    85.4~(0.99)	& 72.4~(0.65)	& 57.8~(0.49)	& 53.4~(0.60)\\
        & XLNet     & 
                    \textbf{94.7~(0.31)}	& 89.4~(0.24)	& 81.2~(0.27)	& \textbf{80.4~(0.34)} &
                    85.8~(0.84)	& 74.3~(1.79)	& 59.0~(1.88)	& 54.9~(2.64)\\
        & SpanBERT  & 
                    93.9~(0.24)	& 88.2~(0.19)	& 79.3~(0.37)	& 78.4~(0.29) &
                    84.4~(0.53)	& 71.6~(1.52)	& 58.6~(0.50)	& 54.4~(1.01)\\
        & DeBERTa   & 
                    94.6~(0.38)	& \textbf{89.8~(0.65)}	& \textbf{81.0~(0.64)}	& 80.2~(0.70) &
                    \textbf{87.3~(0.69)}	& \textbf{76.0~(0.90)}	& \textbf{60.7~(1.53)}	& \textbf{56.7~(1.46)}\\
        \bottomrule
    \end{tabular}
    \caption{Results for intra-sentential parsing with various language models (\textbf{RST-Parseval}). 
    }
    \label{tab:results_intra}
\end{table*}

Table~\ref{tab:result_base:horizontal_concat} shows the results with different pre-trained language models. 
The scores on RST-DT are better than those on Instr-DT. 
This is attributed to the
size of the datasets. Instr-DT is significantly smaller than RST-DT  
while the number of rhetorical relations is larger. 
In fact, standard deviations on Instr-DT are 
larger than those on RST-DT.
However, the tendencies of the experimental results on both datasets are similar. 


The results obtained from paired bootstrap resampling tests\footnote{
In this paper, we set the significance level ($\alpha$) to 0.05.
} between top-down and bottom-up parsers whilst fixing the 
language model show that significant differences are found only in Span and Nuc. for XLNet on RST-DT, and 
Rel. and Full for SpanBERT on Instr-DT, respectively. 

On the other hand, the performance of the parsers varies widely depending on their language model when fixing the parsing strategy.
To investigate the significant differences among parsers,
we performed multiple comparison tests based on the paired bootstrap resampling tests while controlling the false discovery rate \cite{Benjamini1995}.
The results show that DeBERTa significantly outperformed BERT, SpanBERT, and sometimes significantly outperformed RoBERTa and XLNet in top-down parsing. In contrast, it significantly outperformed BERT, RoBERTa, XLNet, and SpanBERT in bottom-up parsing.
RoBERTa and XLNet obtained similar results; they significantly outperformed
BERT, and sometimes significantly outperformed SpanBERT. 
While SpanBERT only significantly outperformed BERT, it sometimes has comparable performance to RoBERTa and XLNet. 
In particular, the bottom-up parser with DeBERTa outperformed Span, Nuc., Rel., and Full scores of the current best parser \cite{zhang-etal-2021-adversarial} by 1.5, 2.5, 1.7, and 1.6 points, respectively.\footnote{
We compared vanilla top-down and bottom-up parsers under the same conditions. We do not try to discuss which could be better as a parser being decorated with new methodologies. 
}
Furthermore, most parsers yield a performance comparable to current state-of-the-art parsers.

We believe that the results have a significant impact in the RST-parsing community. 
Since we built our baseline parsers based on a simple architecture, as described in Section~\ref{sec:parsing}, 
we can conduct more reliable experiments to reveal the effectiveness of newly proposed methods on top of them without any concern regarding the choice of pre-trained language models or parsing strategies.

While the evaluation results demonstrate that we successfully built baseline parsers, they also raise the following questions
for us: (1) Why did DeBERTa, trained with the half-size dataset (85G), outperform RoBERTa, trained with
the most extensive dataset (161G) (2) Why did SpanBERT consistently outperform BERT with significant differences, even though they are trained with the same dataset (16GB).
It is well known that pre-trained language models trained with large-scale datasets boost the performance 
\cite{DBLP:journals/corr/abs-2001-08361}; however, the above results do not necessarily agree with the finding.

We believe that the results may be due to a {\it span-masking} scheme, a common feature between SpanBERT and DeBERTa.
With the span-masking scheme, randomly generated spans consisting of a sequence of tokens with the length up to 5 (for SpanBERT) or 3 (for DeBERTa) are masked, and then the language models are trained to predict the span boundary tokens in the mask.
That is, the span-masking scheme is considered more context-sensitive than the token-masking scheme.
Thus, pre-trained language models with a span-masking scheme are suitable for obtaining vector representations for long text spans consisting of EDUs.
\renewcommand{\arraystretch}{0.85}
\begin{table*}[t]
    \centering
    \small
    \begin{tabular}{llPPPPPPPP}
    \toprule
    & & \multicolumn{4}{c}{\textbf{RST-DT}} & \multicolumn{4}{c}{\textbf{Instr-DT}}\\       
    \cmidrule(lr){3-6}\cmidrule(lr){7-10}
    & & \textbf{Span} & \textbf{Nuc.} & \textbf{Rel.} & \textbf{Full} & \textbf{Span} & \textbf{Nuc.} & \textbf{Rel.} & \textbf{Full} \\
    \midrule 
    \multirow{5}{*}{\rotatebox[origin=c]{90}{Top-down}}
        & BERT      & 
                    65.7~(1.01)	& 45.8~(0.12)	& 32.0~(0.51)	& 31.3~(0.43) &
                    69.5~(0.64)	& 55.8~(0.20)	& 35.8~(0.59)	& 34.6~(0.61)\\
        & RoBERTa   &
                    73.6~(0.17)	& 54.8~(0.90)	& 41.3~(0.65)	& 40.3~(0.57) &
                    78.3~(0.34)	& 66.7~(0.49)	& 49.3~(0.65)	& 47.9~(0.78)\\
        & XLNet     & 
                    73.8~(0.61)	& 55.7~(0.96)	& 41.7~(0.84)	& 40.5~(0.76) &
                    76.4~(0.12)	& 64.3~(0.34)	& 44.3~(1.11)	& 43.2~(1.06)\\
        & SpanBERT  & 
                    72.1~(0.74)	& 52.9~(0.46)	& 38.7~(0.15)	& 37.5~(0.19) &
                    76.5~(0.83)	& 63.8~(0.90)	& 44.2~(0.63)	& 43.5~(0.58)\\
        & DeBERTa   & 
                    \textbf{74.2~(0.48)}	& \textbf{57.1~(0.11)}	& \textbf{42.7~(0.44)}	& \textbf{41.7~(0.54)} &
                    \textbf{79.8~(0.85)}	& \textbf{68.2~(1.16)}	& \textbf{49.4~(1.37)}	& \textbf{48.3~(1.39)}\\
    \midrule
    \multirow{5}{*}{\rotatebox[origin=c]{90}{Bottom-up}} 
        & BERT      & 
                    64.4~(0.86)	& 45.9~(1.29)	& 32.4~(0.74)	& 31.3~(0.75)&
                    71.6~(0.86)	& 57.9~(1.41)	& 38.0~(2.02)	& 36.4~(1.94)\\
        & RoBERTa   & 
                    72.1~(0.11)	& 53.9~(0.81)	& 39.9~(0.59)	& 38.9~(0.77)&
                    76.4~(0.80)	& 64.7~(0.50)	& 48.0~(1.01)	& 47.1~(1.02)\\
        & XLNet     & 
                    71.9~(1.00)	& 54.2~(1.20)	& 40.3~(1.23)	& 39.5~(1.04) &
                    76.8~(1.17)	& 65.1~(0.73)	& 46.4~(1.76)	& 45.2~(1.76)\\
        & SpanBERT  & 
                    71.7~(0.43)	& 53.4~(0.96)	& 39.6~(1.24)	& 38.7~(1.23)&
                    76.2~(1.29)	& 64.1~(1.36)	& 45.6~(1.01)	& 44.7~(0.94)\\
        & DeBERTa   & 
                    \textbf{74.3~(0.65)}	& \textbf{57.2~(1.08)}	& \textbf{42.9~(0.42)}	& \textbf{41.8~(0.55)}&
                    \textbf{80.6~(0.67)}	& \textbf{68.3~(1.44)}	& \textbf{50.2~(0.79)}	& \textbf{48.8~(0.64)}\\
        \bottomrule
    \end{tabular}
    \caption{Results for multi-sentential parsing with various language models (\textbf{RST-Parseval}). 
    }
    \label{tab:results_multi}
\end{table*}
\renewcommand{\arraystretch}{0.85}
\begin{table}[t]
    \centering
    \small
    {\tabcolsep=5pt
    \begin{tabular}{llcccccc}
    \toprule
    & & \multicolumn{3}{c}{\textbf{RST-DT}} & \multicolumn{3}{c}{\textbf{Instr-DT}}\\       
    \cmidrule(lr){3-5}\cmidrule(lr){6-8}
    & & \textbf{N-S} & \textbf{S-N} & \textbf{N-N} & \textbf{N-S} & \textbf{S-N} & \textbf{N-N} \\
    \midrule 
    \multirow{5}{*}{\rotatebox[origin=c]{90}{Top-down}}
        & BERT      & 
                    60.0& 65.0 & 50.7& 45.8	&30.0	&45.6\\  
       & RoBERTa   &
                    67.7&69.8	&60.5	&56.6&	\textbf{46.6}&	57.1\\
        & XLNet     & 
                    68.7	&\textbf{70.8}&	61.1&	57.1&	38.6&	56.3\\
        & SpanBERT  & 
                    67.1&	68.6&	57.7&	56.1&	33.7&	56.1\\
        & DeBERTa   & 
                    \textbf{69.1}	& 69.6	&\textbf{63.1}&	\textbf{58.8}&	37.9	&\textbf{60.3}\\
    \midrule
    \multirow{5}{*}{\rotatebox[origin=c]{90}{Bottom-up}} 
        & BERT      & 
                58.5	& 65.3	&49.1	&45.8 &	31.7 &	48.6\\
        & RoBERTa   & 
                67.7	&\textbf{71.9}&58.5	&55.5	&\textbf{43.3}&	57.4\\
        & XLNet     & 
        67.1&	69.3 &	59.4 &	58.6 &	40.8 &	57.4\\
        & SpanBERT  & 
        66.2	& 69.1 &	59.4 &	55.1 &	39.4 &	55.1\\
        & DeBERTa   & 
        \textbf{69.6}&	70.7&\textbf{61.4}&\textbf{60.7}&	42.1&	\textbf{62.4}\\        
        \bottomrule
    \end{tabular}
    }
    \caption{Results for nuclearity prediction.
    }
    \label{tab:results_nuc}
\end{table}
To discuss the impact of the span-masking scheme in more detail,
we evaluated our parsers in terms of intra- and multi-sentential parsing performance.
Tables \ref{tab:results_intra} and \ref{tab:results_multi} show the results.\footnote{
Note that Standard-Parseval cannot be applied in this setting because leaf nodes in the gold and predicted RST-trees are not necessarily in one-to-one correspondence. See Appendix \ref{app:intra-multi} for more detail.}
From Table \ref{tab:results_intra}, we can see that the tendency of the results is quite different 
from that in Table \ref{tab:result_base:horizontal_concat}; the differences among parsers 
are smaller than those in Table \ref{tab:result_base:horizontal_concat}.
Particularly, the differences among the four methods except for BERT are within 1 point for Full on RST-DT.
Other noteworthy points include that the scores of BERT are close to those of the other methods, and DeBERTa often did not achieve the best scores.
In contrast, Table \ref{tab:results_multi} emphasizes the effectiveness of DeBERTa and SpanBERT. DeBERTa completely outperformed the other methods with large differences, and the differences between SpanBERT and BERT became larger.
To obtain better results in multi-sentential parsing, we need good representations for longer text spans over 
sentences. 
Thus, we believe that the span-masking scheme would help generate better representations for the longer text spans.
The results also reveal that there is still much more room for further improvement than intra-sentential parsing.

Finally, we show another piece of evidence for the effectiveness of the span-masking scheme in Table \ref{tab:results_nuc}, which demonstrates the performance of nuclearity prediction among N-S, S-N, and N-N.
N-N relations originally occur in n-array (n $>$ 2) trees in many cases. Therefore, we need 
good representations for longer text spans to detect N-N relations accurately.
From the table, we can see that DeBERTa achieved the best for N-N with large gains, over 2 points on RST-DT and over 3 points on InstrDT.
Furthermore, SpanBERT is sometimes comparable to XLNet and RoBERTa in this table.
Those results indicate that the span-masking scheme is effective in obtaining good representations for longer text spans again.
The impact of span-masking scheme may lead to novel research perspectives toward RST parsing-specific language models.
    \section{Conclusion}

This paper explored ways to build strong baselines for RST parsing, based on existing top-down and bottom-up parsers,
while varying the use of five transformer-based pre-trained language models: BERT, RoBERTa, XLNet, SpanBERT, and DeBERTa. 
We employed a span-based model as a top-down parser and a shift-reduce model as a bottom-up parser. The experimental results obtained from the RST-DT and Instr-DT revealed that the language models, except for BERT, boost the performance of RST parsing in both strategies. 
The DeBERTa-based bottom-up parser achieved the best scores; in particular, the fully labeled span F1 score of 55.4 on
the RST-DT. 
Furthermore, our experimental results implied that language models with a span-masking scheme, such as SpanBERT, DeBERTa, are suitable for RST parsing since they would generate better representations for long text spans than those with a token-masking scheme.

    \section*{Limitations}

As shown in the experimental results, our approach would not perform well with insufficient training data. For example, the performance obtained from Instr-DT was inferior to that obtained from RST-DT in Rel. and Full. The results were caused by the small amount of training data and many rhetorical relations. Since the annotation costs for RST are considerable, how we obtain enough high-quality data is a significant issue for building RST parsers for new domains and languages. Furthermore, since our parsers rely on a large-scale pre-trained language model, they do not perform well for languages that are not ready to use the pre-trained language model, such as low-resource languages. 
In future work, we should improve the domain/language portability, and we believe the following are practical approaches:  
\begin{enumerate}
	\item Introducing a multilingual pre-trained language model,
	\item Introducing transfer learning and data augmentation.
\end{enumerate}


\section*{Acknowledgements}
Part of this work was supported by JSPS KAKENHI Grant Numbers P21H03505.     
    \bibliography{anthology,custom}

\begin{thebibliography}{41}
\expandafter\ifx\csname natexlab\endcsname\relax\def\natexlab#1{#1}\fi

\bibitem[{Benjamini and Hochberg(1995)}]{Benjamini1995}
Yoav Benjamini and Yosef Hochberg. 1995.
\newblock Controlling the false discovery rate: A practical and powerful
  approach to multiple testing.
\newblock \emph{Journal of the Royal Statistical Society Series B
  (Methodological)}, 57(1):289--300.

\bibitem[{Bhatia et~al.(2015)Bhatia, Ji, and
  Eisenstein}]{bhatia-etal-2015-better}
Parminder Bhatia, Yangfeng Ji, and Jacob Eisenstein. 2015.
\newblock \href {https://doi.org/10.18653/v1/D15-1263} {Better document-level
  sentiment analysis from {RST} discourse parsing}.
\newblock In \emph{Proceedings of the 2015 Conference on Empirical Methods in
  Natural Language Processing}, pages 2212--2218, Lisbon, Portugal. Association
  for Computational Linguistics.

\bibitem[{Denkowski and Neubig(2017)}]{denkowski-neubig-2017-stronger}
Michael Denkowski and Graham Neubig. 2017.
\newblock \href {https://doi.org/10.18653/v1/W17-3203} {Stronger baselines for
  trustable results in neural machine translation}.
\newblock In \emph{Proceedings of the First Workshop on Neural Machine
  Translation}, pages 18--27, Vancouver. Association for Computational
  Linguistics.

\bibitem[{duVerle and Prendinger(2009)}]{duverle-prendinger-2009-novel}
David duVerle and Helmut Prendinger. 2009.
\newblock \href {https://www.aclweb.org/anthology/P09-1075} {A novel discourse
  parser based on support vector machine classification}.
\newblock In \emph{Proceedings of the Joint Conference of the 47th Annual
  Meeting of the {ACL} and the 4th International Joint Conference on Natural
  Language Processing of the {AFNLP}}, pages 665--673, Suntec, Singapore.
  Association for Computational Linguistics.

\bibitem[{{Falcon et al.}(2019)}]{falcon2019pytorch}
William {Falcon et al.} 2019.
\newblock \href {https://github.com/PyTorchLightning/pytorch-lightning}
  {Pytorch lightning}.
\newblock \emph{GitHub. Note:
  https://github.com/PyTorchLightning/pytorch-lightning}, 3.

\bibitem[{Feng and Hirst(2012)}]{feng-hirst-2012-text}
Vanessa~Wei Feng and Graeme Hirst. 2012.
\newblock \href {https://www.aclweb.org/anthology/P12-1007} {Text-level
  discourse parsing with rich linguistic features}.
\newblock In \emph{Proceedings of the 50th Annual Meeting of the Association
  for Computational Linguistics (Volume 1: Long Papers)}, pages 60--68, Jeju
  Island, Korea. Association for Computational Linguistics.

\bibitem[{Feng and Hirst(2014)}]{feng-hirst-2014-linear}
Vanessa~Wei Feng and Graeme Hirst. 2014.
\newblock \href {https://doi.org/10.3115/v1/P14-1048} {A linear-time bottom-up
  discourse parser with constraints and post-editing}.
\newblock In \emph{Proceedings of the 52nd Annual Meeting of the Association
  for Computational Linguistics (Volume 1: Long Papers)}, pages 511--521,
  Baltimore, Maryland. Association for Computational Linguistics.

\bibitem[{Gao et~al.(2020)Gao, Wu, Li, Joty, Hoi, Xiong, King, and
  Lyu}]{gao-etal-2020-discern}
Yifan Gao, Chien-Sheng Wu, Jingjing Li, Shafiq Joty, Steven~C.H. Hoi, Caiming
  Xiong, Irwin King, and Michael Lyu. 2020.
\newblock \href {https://doi.org/10.18653/v1/2020.emnlp-main.191} {Discern:
  Discourse-aware entailment reasoning network for conversational machine
  reading}.
\newblock In \emph{Proceedings of the 2020 Conference on Empirical Methods in
  Natural Language Processing (EMNLP)}, pages 2439--2449, Online. Association
  for Computational Linguistics.

\bibitem[{Guz and Carenini(2020)}]{guz-carenini-2020-coreference}
Grigorii Guz and Giuseppe Carenini. 2020.
\newblock \href {https://doi.org/10.18653/v1/2020.codi-1.17} {Coreference for
  discourse parsing: A neural approach}.
\newblock In \emph{Proceedings of the First Workshop on Computational
  Approaches to Discourse}, pages 160--167, Online. Association for
  Computational Linguistics.

\bibitem[{Guz et~al.(2020)Guz, Huber, and Carenini}]{guz-etal-2020-unleashing}
Grigorii Guz, Patrick Huber, and Giuseppe Carenini. 2020.
\newblock \href {https://doi.org/10.18653/v1/2020.coling-main.337} {Unleashing
  the power of neural discourse parsers - a context and structure aware
  approach using large scale pretraining}.
\newblock In \emph{Proceedings of the 28th International Conference on
  Computational Linguistics}, pages 3794--3805, Barcelona, Spain (Online).
  International Committee on Computational Linguistics.

\bibitem[{He et~al.(2021)He, Liu, Gao, and Chen}]{he2021deberta}
Pengcheng He, Xiaodong Liu, Jianfeng Gao, and Weizhu Chen. 2021.
\newblock \href {https://openreview.net/forum?id=XPZIaotutsD} {{DeBERTa}:
  Decoding-enhanced {BERT} with disentangled attention}.
\newblock In \emph{Proceedings of the International Conference on Learning
  Representations}.

\bibitem[{Heilman and Sagae(2015)}]{heilman2015fast}
Michael Heilman and Kenji Sagae. 2015.
\newblock \href {http://arxiv.org/abs/1505.02425} {Fast rhetorical structure
  theory discourse parsing}.
\newblock \emph{CoRR}, abs/1505.02425.

\bibitem[{Ji and Eisenstein(2014)}]{ji-eisenstein-2014-representation}
Yangfeng Ji and Jacob Eisenstein. 2014.
\newblock \href {https://doi.org/10.3115/v1/P14-1002} {Representation learning
  for text-level discourse parsing}.
\newblock In \emph{Proceedings of the 52nd Annual Meeting of the Association
  for Computational Linguistics (Volume 1: Long Papers)}, pages 13--24,
  Baltimore, Maryland. Association for Computational Linguistics.

\bibitem[{Joty et~al.(2013)Joty, Carenini, Ng, and
  Mehdad}]{joty-etal-2013-combining}
Shafiq Joty, Giuseppe Carenini, Raymond Ng, and Yashar Mehdad. 2013.
\newblock \href {https://www.aclweb.org/anthology/P13-1048} {Combining intra-
  and multi-sentential rhetorical parsing for document-level discourse
  analysis}.
\newblock In \emph{Proceedings of the 51st Annual Meeting of the Association
  for Computational Linguistics (Volume 1: Long Papers)}, pages 486--496,
  Sofia, Bulgaria. Association for Computational Linguistics.

\bibitem[{Joty et~al.(2015)Joty, Carenini, and Ng}]{joty-etal-2015-codra}
Shafiq Joty, Giuseppe Carenini, and Raymond~T. Ng. 2015.
\newblock \href {https://doi.org/10.1162/COLI_a_00226} {{CODRA}: A novel
  discriminative framework for rhetorical analysis}.
\newblock \emph{Computational Linguistics}, 41(3):385--435.

\bibitem[{Kaplan et~al.(2020)Kaplan, McCandlish, Henighan, Brown, Chess, Child,
  Gray, Radford, Wu, and Amodei}]{DBLP:journals/corr/abs-2001-08361}
Jared Kaplan, Sam McCandlish, Tom Henighan, Tom~B. Brown, Benjamin Chess, Rewon
  Child, Scott Gray, Alec Radford, Jeffrey Wu, and Dario Amodei. 2020.
\newblock \href {https://arxiv.org/abs/2001.08361} {Scaling laws for neural
  language models}.
\newblock \emph{CoRR}, abs/2001.08361.

\bibitem[{Kobayashi et~al.(2020)Kobayashi, Hirao, Kamigaito, Okumura, and
  Nagata}]{kobayashi-etal-2020-topdown}
Naoki Kobayashi, Tsutomu Hirao, Hidetaka Kamigaito, Manabu Okumura, and Masaaki
  Nagata. 2020.
\newblock \href {https://doi.org/10.1609/aaai.v34i05.6321} {Top-down rst
  parsing utilizing granularity levels in documents}.
\newblock In \emph{Proceedings of the AAAI Conference on Artificial
  Intelligence}, pages 8099--8106.

\bibitem[{Kobayashi et~al.(2021)Kobayashi, Hirao, Kamigaito, Okumura, and
  Nagata}]{kobayashi-etal-2021-improving}
Naoki Kobayashi, Tsutomu Hirao, Hidetaka Kamigaito, Manabu Okumura, and Masaaki
  Nagata. 2021.
\newblock \href {https://doi.org/10.18653/v1/2021.naacl-main.127} {Improving
  neural {RST} parsing model with silver agreement subtrees}.
\newblock In \emph{Proceedings of the 2021 Conference of the North American
  Chapter of the Association for Computational Linguistics: Human Language
  Technologies}, pages 1600--1612, Online. Association for Computational
  Linguistics.

\bibitem[{Koto et~al.(2021)Koto, Lau, and Baldwin}]{koto-etal-2021-top}
Fajri Koto, Jey~Han Lau, and Timothy Baldwin. 2021.
\newblock \href {https://www.aclweb.org/anthology/2021.eacl-main.60} {Top-down
  discourse parsing via sequence labelling}.
\newblock In \emph{Proceedings of the 16th Conference of the European Chapter
  of the Association for Computational Linguistics: Main Volume}, pages
  715--726, Online. Association for Computational Linguistics.

\bibitem[{Li et~al.(2014)Li, Li, and Hovy}]{li-etal-2014-recursive}
Jiwei Li, Rumeng Li, and Eduard Hovy. 2014.
\newblock \href {https://doi.org/10.3115/v1/D14-1220} {Recursive deep models
  for discourse parsing}.
\newblock In \emph{Proceedings of the 2014 Conference on Empirical Methods in
  Natural Language Processing ({EMNLP})}, pages 2061--2069, Doha, Qatar.
  Association for Computational Linguistics.

\bibitem[{Li et~al.(2016)Li, Li, and Chang}]{li-etal-2016-discourse}
Qi~Li, Tianshi Li, and Baobao Chang. 2016.
\newblock \href {https://doi.org/10.18653/v1/D16-1035} {Discourse parsing with
  attention-based hierarchical neural networks}.
\newblock In \emph{Proceedings of the 2016 Conference on Empirical Methods in
  Natural Language Processing}, pages 362--371, Austin, Texas. Association for
  Computational Linguistics.

\bibitem[{Liang et~al.(2019)Liang, Lim, Tsai, Salakhutdinov, and
  Morency}]{liang-etal-2019-strong}
Paul~Pu Liang, Yao~Chong Lim, Yao-Hung~Hubert Tsai, Ruslan Salakhutdinov, and
  Louis-Philippe Morency. 2019.
\newblock \href {https://doi.org/10.18653/v1/N19-1267} {Strong and simple
  baselines for multimodal utterance embeddings}.
\newblock In \emph{Proceedings of the 2019 Conference of the North {A}merican
  Chapter of the Association for Computational Linguistics: Human Language
  Technologies, Volume 1 (Long and Short Papers)}, pages 2599--2609,
  Minneapolis, Minnesota. Association for Computational Linguistics.

\bibitem[{Lin et~al.(2019)Lin, Joty, Jwalapuram, and
  Bari}]{lin-etal-2019-unified}
Xiang Lin, Shafiq Joty, Prathyusha Jwalapuram, and M~Saiful Bari. 2019.
\newblock \href {https://doi.org/10.18653/v1/P19-1410} {A unified linear-time
  framework for sentence-level discourse parsing}.
\newblock In \emph{Proceedings of the 57th Annual Meeting of the Association
  for Computational Linguistics}, pages 4190--4200, Florence, Italy.
  Association for Computational Linguistics.

\bibitem[{Liu and Chen(2019)}]{liu-chen-2019-exploiting}
Zhengyuan Liu and Nancy Chen. 2019.
\newblock \href {https://doi.org/10.18653/v1/D19-5415} {Exploiting
  discourse-level segmentation for extractive summarization}.
\newblock In \emph{Proceedings of the 2nd Workshop on New Frontiers in
  Summarization}, pages 116--121, Hong Kong, China. Association for
  Computational Linguistics.

\bibitem[{Loshchilov and Hutter(2017)}]{loshchilov2019decoupled}
Ilya Loshchilov and Frank Hutter. 2017.
\newblock \href {http://arxiv.org/abs/1711.05101} {Fixing weight decay
  regularization in adam}.
\newblock \emph{CoRR}, abs/1711.05101.

\bibitem[{Lynn~Carlson(2002)}]{carlson-etal-2002-rstdt}
Mary Ellen~Okurowski Lynn~Carlson, Daniel~Marcu. 2002.
\newblock \href {https://catalog.ldc.upenn.edu/LDC2002T07} {\emph{RST Discourse
  Treebank}}.
\newblock Philadelphia: Linguistic Data Consortium.

\bibitem[{Mann and Thompson(1987)}]{mann:87:a}
W.C. Mann and S.A Thompson. 1987.
\newblock Rhetorical structure theory: A theory of text organization.
\newblock Technical Report ISI/RS-87-190, USC/ISI.

\bibitem[{Morey et~al.(2017)Morey, Muller, and Asher}]{morey-etal-2017-much}
Mathieu Morey, Philippe Muller, and Nicholas Asher. 2017.
\newblock \href {https://doi.org/10.18653/v1/D17-1136} {How much progress have
  we made on {RST} discourse parsing? a replication study of recent results on
  the {RST}-{DT}}.
\newblock In \emph{Proceedings of the 2017 Conference on Empirical Methods in
  Natural Language Processing}, pages 1319--1324, Copenhagen, Denmark.
  Association for Computational Linguistics.

\bibitem[{Nguyen et~al.(2021)Nguyen, Nguyen, Joty, and
  Li}]{nguyen-etal-2021-rst}
Thanh-Tung Nguyen, Xuan-Phi Nguyen, Shafiq Joty, and Xiaoli Li. 2021.
\newblock \href {https://doi.org/10.18653/v1/2021.naacl-main.128} {{RST}
  parsing from scratch}.
\newblock In \emph{Proceedings of the 2021 Conference of the North American
  Chapter of the Association for Computational Linguistics: Human Language
  Technologies}, pages 1613--1625, Online. Association for Computational
  Linguistics.

\bibitem[{Paszke et~al.(2019)Paszke, Gross, Massa, Lerer, Bradbury, Chanan,
  Killeen, Lin, Gimelshein, Antiga, Desmaison, Kopf, Yang, DeVito, Raison,
  Tejani, Chilamkurthy, Steiner, Fang, Bai, and Chintala}]{NEURIPS2019_9015}
Adam Paszke, Sam Gross, Francisco Massa, Adam Lerer, James Bradbury, Gregory
  Chanan, Trevor Killeen, Zeming Lin, Natalia Gimelshein, Luca Antiga, Alban
  Desmaison, Andreas Kopf, Edward Yang, Zachary DeVito, Martin Raison, Alykhan
  Tejani, Sasank Chilamkurthy, Benoit Steiner, Lu~Fang, Junjie Bai, and Soumith
  Chintala. 2019.
\newblock \href
  {http://papers.neurips.cc/paper/9015-pytorch-an-imperative-style-high-performance-deep-learning-library.pdf}
  {Pytorch: An imperative style, high-performance deep learning library}.
\newblock In H.~Wallach, H.~Larochelle, A.~Beygelzimer, F.~d\textquotesingle
  Alch\'{e}-Buc, E.~Fox, and R.~Garnett, editors, \emph{Advances in Neural
  Information Processing Systems 32}, pages 8024--8035. Curran Associates, Inc.

\bibitem[{Sagae and Lavie(2005)}]{sagae-lavie-2005-classifier}
Kenji Sagae and Alon Lavie. 2005.
\newblock \href {https://www.aclweb.org/anthology/W05-1513} {A classifier-based
  parser with linear run-time complexity}.
\newblock In \emph{Proceedings of the Ninth International Workshop on Parsing
  Technology}, pages 125--132, Vancouver, British Columbia. Association for
  Computational Linguistics.

\bibitem[{Shi et~al.(2020)Shi, Liu, and Chen}]{shi2020endtoend}
Ke~Shi, Zhengyuan Liu, and Nancy~F. Chen. 2020.
\newblock \href {http://arxiv.org/abs/2012.11169} {An end-to-end document-level
  neural discourse parser exploiting multi-granularity representations}.
\newblock \emph{CoRR}, abs/2012.11169.

\bibitem[{Subba and Di~Eugenio(2009)}]{subba-di-eugenio-2009-effective}
Rajen Subba and Barbara Di~Eugenio. 2009.
\newblock \href {https://www.aclweb.org/anthology/N09-1064} {An effective
  discourse parser that uses rich linguistic information}.
\newblock In \emph{Proceedings of Human Language Technologies: The 2009 Annual
  Conference of the North {A}merican Chapter of the Association for
  Computational Linguistics}, pages 566--574, Boulder, Colorado. Association
  for Computational Linguistics.

\bibitem[{Suzuki et~al.(2018)Suzuki, Takase, Kamigaito, Morishita, and
  Nagata}]{suzuki-etal-2018-empirical}
Jun Suzuki, Sho Takase, Hidetaka Kamigaito, Makoto Morishita, and Masaaki
  Nagata. 2018.
\newblock \href {https://doi.org/10.18653/v1/P18-2097} {An empirical study of
  building a strong baseline for constituency parsing}.
\newblock In \emph{Proceedings of the 56th Annual Meeting of the Association
  for Computational Linguistics (Volume 2: Short Papers)}, pages 612--618,
  Melbourne, Australia. Association for Computational Linguistics.

\bibitem[{Wang et~al.(2017)Wang, Li, and Wang}]{wang-etal-2017-two}
Yizhong Wang, Sujian Li, and Houfeng Wang. 2017.
\newblock \href {https://doi.org/10.18653/v1/P17-2029} {A two-stage parsing
  method for text-level discourse analysis}.
\newblock In \emph{Proceedings of the 55th Annual Meeting of the Association
  for Computational Linguistics (Volume 2: Short Papers)}, pages 184--188,
  Vancouver, Canada. Association for Computational Linguistics.

\bibitem[{Wolf et~al.(2020)Wolf, Debut, Sanh, Chaumond, Delangue, Moi, Cistac,
  Rault, Louf, Funtowicz, Davison, Shleifer, von Platen, Ma, Jernite, Plu, Xu,
  Le~Scao, Gugger, Drame, Lhoest, and Rush}]{wolf-etal-2020-transformers}
Thomas Wolf, Lysandre Debut, Victor Sanh, Julien Chaumond, Clement Delangue,
  Anthony Moi, Pierric Cistac, Tim Rault, Remi Louf, Morgan Funtowicz, Joe
  Davison, Sam Shleifer, Patrick von Platen, Clara Ma, Yacine Jernite, Julien
  Plu, Canwen Xu, Teven Le~Scao, Sylvain Gugger, Mariama Drame, Quentin Lhoest,
  and Alexander Rush. 2020.
\newblock \href {https://doi.org/10.18653/v1/2020.emnlp-demos.6} {Transformers:
  State-of-the-art natural language processing}.
\newblock In \emph{Proceedings of the 2020 Conference on Empirical Methods in
  Natural Language Processing: System Demonstrations}, pages 38--45, Online.
  Association for Computational Linguistics.

\bibitem[{Xu et~al.(2020)Xu, Gan, Cheng, and Liu}]{xu-etal-2020-discourse}
Jiacheng Xu, Zhe Gan, Yu~Cheng, and Jingjing Liu. 2020.
\newblock \href {https://doi.org/10.18653/v1/2020.acl-main.451}
  {Discourse-aware neural extractive text summarization}.
\newblock In \emph{Proceedings of the 58th Annual Meeting of the Association
  for Computational Linguistics}, pages 5021--5031, Online. Association for
  Computational Linguistics.

\bibitem[{Yang et~al.(2019)Yang, Dai, Yang, Carbonell, Salakhutdinov, and
  Le}]{zhilin2019xlnet}
Zhilin Yang, Zihang Dai, Yiming Yang, Jaime Carbonell, Russ~R Salakhutdinov,
  and Quoc~V Le. 2019.
\newblock \href
  {https://proceedings.neurips.cc/paper/2019/file/dc6a7e655d7e5840e66733e9ee67cc69-Paper.pdf}
  {{XLN}et: Generalized autoregressive pretraining for language understanding}.
\newblock In \emph{Advances in Neural Information Processing Systems},
  volume~32. Curran Associates, Inc.

\bibitem[{Yu et~al.(2018)Yu, Zhang, and Fu}]{yu-etal-2018-transition}
Nan Yu, Meishan Zhang, and Guohong Fu. 2018.
\newblock \href {https://www.aclweb.org/anthology/C18-1047} {Transition-based
  neural {RST} parsing with implicit syntax features}.
\newblock In \emph{Proceedings of the 27th International Conference on
  Computational Linguistics}, pages 559--570, Santa Fe, New Mexico, USA.
  Association for Computational Linguistics.

\bibitem[{Zhang et~al.(2021)Zhang, Kong, and
  Zhou}]{zhang-etal-2021-adversarial}
Longyin Zhang, Fang Kong, and Guodong Zhou. 2021.
\newblock \href {https://doi.org/10.18653/v1/2021.acl-long.305} {Adversarial
  learning for discourse rhetorical structure parsing}.
\newblock In \emph{Proceedings of the 59th Annual Meeting of the Association
  for Computational Linguistics and the 11th International Joint Conference on
  Natural Language Processing (Volume 1: Long Papers)}, pages 3946--3957,
  Online. Association for Computational Linguistics.

\bibitem[{Zhang et~al.(2020)Zhang, Xing, Kong, Li, and
  Zhou}]{zhang-etal-2020-top}
Longyin Zhang, Yuqing Xing, Fang Kong, Peifeng Li, and Guodong Zhou. 2020.
\newblock \href {https://doi.org/10.18653/v1/2020.acl-main.569} {A top-down
  neural architecture towards text-level parsing of discourse rhetorical
  structure}.
\newblock In \emph{Proceedings of the 58th Annual Meeting of the Association
  for Computational Linguistics}, pages 6386--6395, Online. Association for
  Computational Linguistics.

\end{thebibliography}
    \bibliographystyle{acl_natbib}
    
    \clearpage 
    \appendix

\section{Hyperparameters \label{app:hyperparames}}
Table \ref{tab:hyperparams} shows the hyperparameters utilized in our experiments.
\begin{table}[t]
    \small
    \centering
    \begin{tabular}{cc}
        \toprule
        {\bf Computing Infrastructure} & GeForce RTX 3090 \\
        \midrule
        \toprule
        \multicolumn{2}{c}{\bf Hyperparameters} \\
        \midrule
        number of training epochs & 20 \\
        \midrule
        patience of early stopping & 5 \\
        \midrule
        batch size & \multirow{2}{*}{5} \\
        (\# of spans/actions) \\
        \midrule
        language model's hidden size & 768 (base) \\
        \midrule
        FFN's hidden size & 512 \\
        \midrule
        dropout & 0.2 \\
        \midrule
        learning rate scheduler & Linear warm-up \\
        \midrule
        optimizer & AdamW \\
        \midrule
        learning rate & {2e-4, 1e-5} \\
        \midrule
        weight decay & 0.01 \\
        \midrule
        gradient clipping & 1.0 \\
        \midrule
        validation criteria & Standard-Parseval: Full \\
        \bottomrule
    \end{tabular}
    \caption{Parameter search space in our experiments.}
    \label{tab:hyperparams}
\end{table}


\section{Evaluation in Intra- and Multi-sentential Parsing \label{app:intra-multi}}
\begin{figure*}[t]
    \centering
    \includegraphics[width=0.9\linewidth]{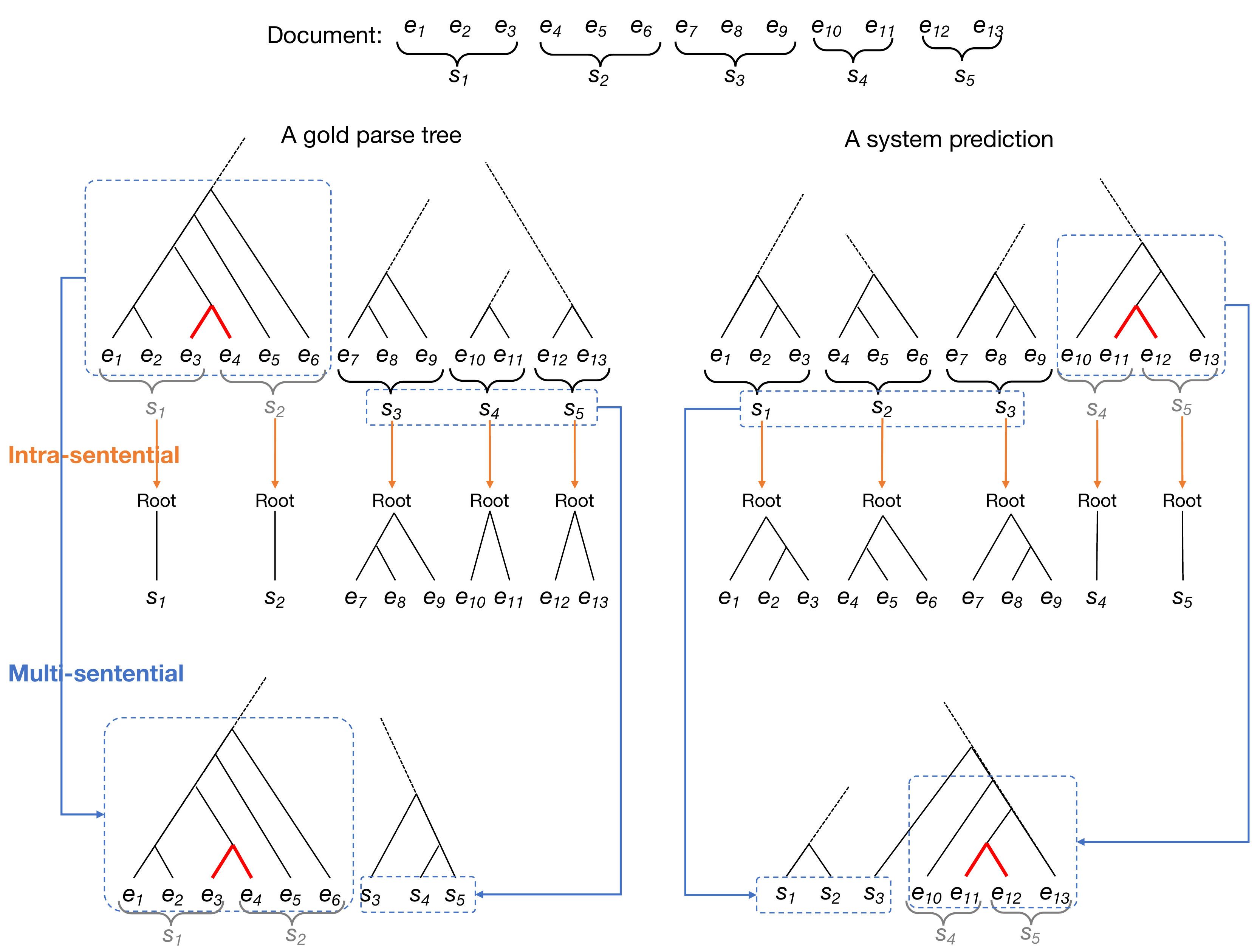}
    \caption{Example of decomposing a RST tree into intra- and multi-sentential trees.}
    \label{fig:intra-multi}
\end{figure*}

Because human annotators sometimes build RST-trees with disregarding sentence boundaries, some RST-trees have span boundaries that disagree with span boundaries in sentences. Figure \ref{fig:intra-multi} shows an example. In the gold tree, 
the subtree consisting of $e_3$ and $e_4$ straddles the boundary between $s_1$ and $s_2$. 
Parsers also sometimes disregard sentence boundaries when building RST-trees.   
In the predicted tree, 
the subtree consisting of $e_{11}$ and $e_{12}$ straddles the boundary between $s_4$ and $s_5$. 
Therefore, we make a best effort to find sentences in the parse trees.

We extract subtrees whose root nodes correspond to sentences when evaluating a predicted tree in terms of intra-sentential parsing. 
In the example, we extract the subtrees corresponding to $s_3$, $s_4$, and $s_5$ from the gold tree and $s_1$, $s_2$, and $s_3$ from the predicted tree. 
However, in this case, $s_1$ and $s_2$ are ignored even though they form valid subtrees in the predicted tree. 
So we give a unary tree whose leaf node is a sentence for $s_1$ and $s_2$ for the gold tree.
Similarly, we give a unary tree for $s_4$ and $s_5$ for the predicted tree (see the middle row in Figure \ref{fig:intra-multi}).
As a result, the leaf nodes of a gold RST-tree do not necessarily have a one-to-one correspondence with those of 
a predicted tree. Thus, we apply RST-Parseval to evaluate predicted trees in terms of intra-sentential parsing.

We replace subtrees corresponding to sentences as leaf nodes when evaluating a predicted tree in multi-sentential parsing. 
In the example, the subtrees dominating $e_7$ to $e_9$, $e_{10}$ to $e_{11}$, and $e_{12}$ to $e_{13}$ in the gold tree are respectively replaced with the leaf nodes $s_3$, $s_4$, and $s_5$. 
Since the gold RST-tree does not have valid subtrees dominating $e_1$ to $e_3$ and $e_4$ to $e_6$, 
we do not replace them as $s_1$ and $s_2$, respectively.
That is, subtrees that cannot be converted into leaf nodes as sentences are left as they are.
Similarly, the subtrees dominating $e_1$ to $e_3$ and $e_4$ to $e_6$ in the predicted tree are respectively replaced as leaf nodes $s_1$ and $s_2$.
We also do not replace $e_{10}$ to $e_{11}$ and $e_{12}$ to $e_{13}$ as $s_4$ and $s_5$ 
(see the bottom row in Figure \ref{fig:intra-multi}).
The transformation may break down the one-to-one correspondence between leaf nodes of gold and predicted RST-trees.
Thus, we also apply RST-Parseval to evaluate predicted trees in terms of multi-sentential parsing.

\end{document}